\pdfoutput=1
\documentclass[11pt, preprint]{article}
\PassOptionsToPackage{dvipsnames,table}{xcolor}
\usepackage[preprint]{acl}
\usepackage{tcolorbox}
\usepackage{enumitem}

\usepackage{times}
\usepackage{latexsym}
\usepackage{float}
\usepackage{array}
\usepackage{graphicx}
\usepackage{enumitem}
\usepackage{tablefootnote}
\usepackage{adjustbox}
\usepackage{stfloats}
\usepackage{listings}
\usepackage{tabularx}
\newcolumntype{L}{X}
\newcolumntype{C}{>{\centering\arraybackslash}X}
\newcolumntype{R}{>{\raggedleft\arraybackslash}X}

\usepackage{lipsum}

\newcommand\nds[1]{\textbf{\textcolor{red}{NdS: #1}}}

\usepackage[T1]{fontenc}
\usepackage{placeins}
\usepackage{makecell}
\usepackage{pifont}
\newcommand{\tick}{\textcolor{Green}{\ding{51}}}
\newcommand{\cross}{\textcolor{Red}{\ding{55}}}

\usepackage[utf8]{inputenc}

\usepackage{microtype}

\usepackage{inconsolata}
\usepackage{xcolor}
\usepackage{colortbl}
\usepackage{hhline}
\usepackage{graphicx}
\usepackage{multirow}
\usepackage{threeparttable}

\usepackage{natbib}
\usepackage{amsmath}
\newcolumntype{g}{>{\columncolor[gray]{0.9}}c}
\usepackage{placeins}
\newcommand{\AZ}{\texttt{az}} %Azerbaijani
\newcommand{\BG}{\texttt{bg}} %Bulgarian
\newcommand{\FR}{\texttt{fr}} %French
\newcommand{\HI}{\texttt{hi}} %Hindi
\newcommand{\KO}{\texttt{ko}} %Korean
\newcommand{\SI}{\texttt{si}} %Sinhala
\newcommand{\VI}{\texttt{vi}} %Vietnamese
\newcommand{\ID}{\texttt{id}} %Indonesian
\newcommand{\TK}{\texttt{tk}} %Turkish

\newcommand{\LMS}{\texttt{LMSpell}}

\title{\LMS: Spell Correction with Pre-Trained Language Models}

\author{
  \ Akesh Gunathilake\textsuperscript{a\thanks{Joint First authors}},
  \ Nadil Karunarathna\textsuperscript{a}\footnotemark[1],
  \ Tharusha Bandaranayake\textsuperscript{a}\footnotemark[1],\\
  \ \textbf{Nisansa de Silva}\textsuperscript{a},
  \ \textbf{Surangika Ranathunga}\textsuperscript{b}
  \ \textbf{Nevidu Jayatilleke}\textsuperscript{a}
  \\
  \textsuperscript{a}Department of Computer Science and Engineering, University of
Moratuwa, Katubedda, 10400, Sri Lanka\\
  \texttt{\{akesh.20,nadil.20,tharusha.20,NisansaDdS,nevidu.25\}@cse.mrt.ac.lk},\\
  \textsuperscript{b}School of Mathematical and Computational Sciences, Massey
University, Auckland, 102904, New Zealand\\
\texttt{s.ranathunga@massey.ac.nz}, \\
}

\begin{document}
\maketitle
\begin{abstract}
%This document is a supplement to the general instructions for *ACL authors. It contains instructions for using the \LaTeX{} style files for ACL conferences.
%The document itself conforms to its own specifications, and is therefore an example of what your manuscript should look like. These instructions should be used both for papers submitted for review and for final versions of accepted papers.
Spell correction is still a challenging problem for low-resource languages (LRLs). While pretrained language models (PLMs) have been employed for spell correction, their use is still limited to a handful of languages, and there has been no proper comparison across PLMs. We present the first empirical study on the effectiveness of PLMs for spell correction, which includes LRLs. We find that Large Language Models (LLMs) outperform their counterparts (encoder-based and encoder-decoder) when the fine-tuning dataset is large. This observation holds even in languages for which the LLM is not pre-trained. We release LMSpell, an easy-to-use spell correction toolkit across PLMs. It includes an evaluation function that compensates for the hallucination of LLMs. Further, we present a case study with Sinhala to shed light on the plight of spell correction for LRLs.
\end{abstract}

\section{Introduction}

Human-written text is prone to spelling errors~\cite{flor-futagi-2012-using}, and not having spell correction tools could adversely impact other Natural Language Processing (NLP) tasks. %NLP systems trained with digitized text containing spelling errors are predisposed to subpar performance. 
For example,~\citet{sonnadara2021sinhala} showed that spell correction has a positive impact on text classification. Training Large Language Models (LLMs) with text containing spelling errors could result in them generating text with spelling errors. Given the current trend of users taking LLM output as undisputed fact~\cite{karunaratne2023new}, this could initiate a vicious cycle that reduces the spelling knowledge of humans who use such generated text.

For high-resource languages (HRLs) such as English, spell correction is more or less a solved problem, albeit somewhat through the brute-forced sheer momentum of the abundance of data, rather than pure innovation; a luxury that low-resource languages (LRLs) do not possess. For example, for the LRL Sinhala, there exists only a handful of research on spell correction, and their accuracy is not par for practical use~\cite{de2019survey}. 
\begin{table}
\tiny  %\tick\cross
\renewcommand{\arraystretch}{1.0}
  \centering
  \begin{tabularx}{0.48\textwidth}{|l|l|C|C|C|}
    \hline
    \textbf{Paper} & \textbf{Language} & \multicolumn{3}{c|}{\textbf{Architectures Compared}} \\
    \hhline{~~---}
    & & \textbf{EO} & \textbf{DO} & \textbf{ED} \\
    \hline
    \citet{9160935} & English & \tick & \tick & \tick \\
    \citet{liu2024chinesespellingcorrectionrephrasing} & Chinese & \tick & \tick & \cross \\
    \citet{martynov-etal-2024-methodology} & Russian, English & \tick & \cross & \tick \\
    \citet{su2024ucsc} & Chinese  & \tick & \cross & \tick \\
    \citet{jiang-etal-2024-chinese} & Chinese & \tick & \tick & \cross \\
    \hline
  \end{tabularx}
  \caption{Empirical studies on PLMs for spell correction. EO - Encoder-Only, DO - Decoder-Only, and ED - Encoder-Decoder}
  \label{tab:Different Architectures}
\end{table}

With their ability to consider contextual information, Deep Learning (DL) techniques have been used for spell correction. The earlier common use of recurrent models~\cite{sakaguchi2017robsut} have yielded center stage today to pre-trained language models (PLMs), which have shown to outperform off-the-shelf spell correctors such as HunSpell and JamSpell by a significant margin~\cite{jayanthi-etal-2020-neuspell, martynov-etal-2024-methodology}.~\citet{sonnadara2021sinhala} showed that a HunSpell based spell corrector is far worse even than recurrent models for Sinhala.~\citet{jayanthi-etal-2020-neuspell} posit that the reason for this is the inability of tools such as HunSpell to utilize contextual information. 

According to Table \ref{tab:Different Architectures}, only~\citet{9160935} experimented with all three PLM types (encoder-based, decoder-based (aka LLMs)
, and encoder-decoder) for spell correction. But they considered only English. So far, only three languages - English, Chinese, and Russian - all HRLs, have been considered in such comparative studies. 
Table \ref{tab:PLM Research} contains a (nearly) comprehensive list of research that employed PLMs for spell correction. 47\% focused on Chinese and 18\% on English, both are HRLs~\citep{ranathunga-de-silva-2022-languages}. In fact, only 7 studies worked on LRLs that fall into Category 3 or lower (which is fewer than the number of studies conducted on English). From the work on LRLs Sinhala~\cite{sudesh2022erroff, sonnadara2021sinhala} and Bengali~\cite{rahman-etal-2023-bspell,banik2024bert} stand out as the languages for which, multiple attempts have been made. A majority of the efforts seem to be using BERT or its various derivatives.
\begin{table}[H]
\tiny
  \centering
  \resizebox{0.48\textwidth}{!}{
  \begin{tabular}{
    |>{\raggedright\arraybackslash}p{2.25cm}
    |>{\raggedright\arraybackslash}p{1.3cm}
    |>{\raggedright\arraybackslash}p{3.3cm}|
  }
    \hline
    \textbf{Paper} & \textbf{Language} & \textbf{Models Used} \\
    \hline
    % 2020  
    \citet{cheng-etal-2020-spellgcn} & Chinese (5) & BERT \\    
    \citet{zhang-etal-2020-spelling} & Chinese (5) & BERT \\
    \citet{9160935} & English (5) & GloVe, fastText, ELMo, GPT, GPT-2, BERT, RoBERTa, XLM-RoBERTa, BART, T5, XLNet \\
    \citet{hu2020misspelling} & English (5) & BERT \\
    \citet{jayanthi-etal-2020-neuspell} & Sinhala (2)  & BERT \\
    \citet{liu-etal-2021-plome} & Chinese (5) & BERT \\
    % 2022
    \citet{tohidian2022bedspell} & English (5) & BERT \\ 
    \citet{ngo2022combination} & Vietnamese (4) & BERT \\
    \citet{mitreska2022nlp} & Croatian (3) & mBERT, DistilBERT, XLM-RoBERTa \\
    \citet{sudesh2022erroff} & Sinhala (2) & mT5, mBART \\
    % 2023    
    \citet{rahman-etal-2023-bspell} & Bengali (3) & BERT \\    
    \citet{wei-etal-2023-ptcspell} & Chinese (5) & BERT \\
    \citet{wu-etal-2023-rethinking} & Chinese (5) & BERT \\
    \citet{huang-etal-2023-frustratingly} & Chinese (5) & BERT \\
    \citet{zhang2023does} & Chinese (5) & GPT-3.5-Turbo \\
    \citet{li2023effectiveness} & Chinese (5) & BERT, BART, BAICHUAN-13B-Chat. Text-Davinci-GPT-3.5 \\ 
    \citet{pankam2023two} & Thai (3) & WangchanBERTa \\
    \citet{ramaneedi2023kannada} & Kannada (2) & mT5 \\
    % 2024
    \citet{martynov-etal-2024-methodology} & Russian (4), English (5) & BERT, T5, GPT-3.5, GPT-4.0, M2M-100 \\
    \citet{su2024ucsc} & Chinese (5) & BERT, BART \\
    \citet{wang2024local} & Chinese (5) & BERT \\
    \citet{liu2024chinesespellingcorrectionrephrasing} & Chinese (5) & BERT, BAICHUAN, GPT2 \\  
    \citet{wu-etal-2024-bi} & Chinese (5) & BERT, ChineseBERT \\
    \citet{jiang-etal-2024-chinese} & Chinese (5) & BERT, GPT-3.5, BAICHUAN \\
    \citet{li-etal-2024-c} & Chinese (5) & BERT, C-LLM, GPT-4,  \\
    \citet{dutta2024enhancing} & English (5) & BART, T5 \\
    \citet{wen2024english} & English (5) & BERT \\
    \citet{guo-etal-2024-retrieval} & English (5) & BERT, Mistral-7B, Claude-3-sonnet \\
    \citet{naziri2024comprehensive} & Persian (4) & BERT \\
    \citet{10837900} & Turkish (4) & TURNA \\    
    \citet{banik2024bert} & Bengali (3) & BERT \\    
    \citet{zhou-etal-2024-simple} & Chinese (5) & BERT, BAICHUAN2, Qwen2.5, InternLM2.5 % new 
    \\
    % 2025
    
    \citet{dong-etal-2025-retrieval} & Chinese (5) & BERT, GPT-3.5, GLM4 \\
    \citet{behera2025hilearners} & Hindi (3) & IndicBert, mT5, mBart \\
    \citet{qiao2025mixture} & Chinese (5) & BERT, BAICHUAN2, Qwen2.5, InternL \\
    \citet{turhan2025leveraging} & Turkish (4) & BERT, XLM-RoBERTa, T5 \\
    \citet{11206935} & Turkish (4) & DistilBERT \\
    \citet{11100636} & Indonesian (4) & T5 Base, Bart \\
    \citet{app16052440} & Chinese (5) & RoBERTa, BERT-wwm, MacBERT \\
   \hline
  \end{tabular}
  }
  \caption{Research that used PLMs for spell correction. The~\citet{ranathunga-de-silva-2022-languages} language category is given in parentheses after the name of each language.}
  \label{tab:PLM Research}
\end{table}

% toatl - 38
% English - 7 = 18%
% Chinese - 18 = 47%
% LRL - 7
% Encoder Models - 33

A hindrance to the use of such PLMs for spell correction is the lack of an easy-to-use library. The \texttt{NeuSpell} library~\cite{jayanthi-etal-2020-neuspell}, which is ostensibly a general framework to implement spell correctors supports only BERT and its derivatives. Regardless, \texttt{NeuSpell} is now nearly deprecated due to lack of frequent updates. 

In response, we present \LMS{}\footnote{\url{https://github.com/Nadil-K/lm-spell}}(CC BY 4.0), which is an easy-to-use spell correction toolkit that can be extended for any language and any type of PLM. We also introduce a new script for evaluating spell correction, which takes hallucination into consideration. In this research, we conduct a comprehensive evaluation of different PLMs in the context of a range of languages, including LRLs. We empirically show that even a small LLM (Gemma 3 270M) or an encoder-decoder PLM (mBART50) fine-tuned with just 5k sentences is sufficient to beat rule-based JamSpell and ASpell. For LLMs, even creating this 5k dataset with synthetic spelling errors is sufficient to beat the rule-based systems, which further strengthens the possibility of pragmatically using LLMs for the task of spell correction. When comparing performance across different PLM types, we note that mBART still has an advantage over Gemma and Llama, at comparable parameter counts. However, larger LLMs have a clear advantage over mBART. LLMs also demonstrate a better capacity at learning - when trained with larger dataset sizes, LLMs (even smaller Gemma 270M) outperform their encoder-based and encoder-decoder counterparts for Sinhala. Our case study on Sinhala spell correction demonstrates possible improvements to spell correction with LLMs, as well as the robustness of LLMs for the task, in the context of LRLs.

\section{Methodology}
\LMS{} supports the implementation of spell correctors using all three types of PLMs (encoder-based, decoder-based, and encoder-decoder). It abstracts basic PLM-related functions such as fine-tuning and inference, allowing users to select a model without needing to consider model-specific implementation details. Internally, the relevant classes are dynamically loaded based on the chosen model.  

%How each PLM architecture was employed for the task of spell correction is discussed below. 

\paragraph{Encoder-based:} Encoder-based PLMs have been adapted for spell correction by framing it as a sequence labelling task~\cite{alyafeai2020survey}.  However, most research on encoder-only models focused either on error detection~\cite{jayanthi-etal-2020-neuspell} or error correction~\cite{zhang-etal-2020-spelling, banik2024bert}, while a different technique (e.g., a dictionary-based approach) was used for the other task. In order to implement both error detection and correction only using an encoder-based model, we follow~\citet{hong2019faspell} and implement sequence labeling such that each input token is assigned a corresponding output token. If the token is correct, it is labeled with itself, whereas an erroneous token is labeled with its corrected form. This was achieved using masked language modeling (MLM). We fine-tuned the PLM to treat every token as a masked token, thereby enabling both error detection and correction within the encoder-based architecture.

\paragraph{Decoder-based:} Modern-day LLMs produce an output in response to the task specified as a textual prompt. Our prompt is in Appendix~\ref{sec:lmspell}. It gets adjusted for the user's choice of language (lang). 

\paragraph{Encoder-decoder-based:} The PLM is fine-tuned to generate the spell-corrected sentence, when an erroneous sentence input is given as a seq-to-seq task. Some PLMs in this category expect a special token before the input sequence to indicate the language, while some infer the language automatically. Both cases are supported. Further implementation details are in Appendix~\ref{sec:lmspell}.

\section{Experiment Setup}
\label{sec:experiment_setup}
We experimented with the following PLMs: encoder-based (XLM-R), encoder-decoder (mT5, mBART50) and the Instruct version of the decoder-based LLMs (Llama 3.1 8B, Llama 3.2 1B, Gemma 2 9B, Gemma 3 1B,  Gemma 3 270m). All LLMs, except Gemma 3 270m were fine-tuned using LoRA. Furthermore, stated results are taken from 1 epoch for all LLMs. We experimented with Azerbaijani (\AZ), Bulgarian (\BG), French (\FR), Hindi (\HI), Korean (\KO), Sinhala (\SI), Vietnamese (\VI), Indonesian (\ID), and Turkish(\TK). Language selection was based on public dataset availability. Note that the character-level edits of the \HI\ dataset are focused on fixing grammar issues. Nevertheless, we used this dataset, because this is the second largest training dataset we could find at the time. Details of the PLMs and the languages are in Tables~\ref{tab: PLMs}, and~\ref{tab:languages} (respectively). For additional details see Appendices~\ref{sec:hyper-params} and~\ref{app:lang}. Hyper-parameter details in Tables~\ref{tab:training_parameters_1} and~\ref{tab:training_parameters_2}. Computational cost is given in Appendix~\ref{sec:computational_resources}. \text{F0.5-score}\footnote{F0.5 over F1 is favored - missing an erroneous word is better than making an erroneous correction~\cite{ng-etal-2014-conll}. } is used to evaluate the correction task. F1-score is used to evaluate the detection task.

\begin{table}[h!tb]
\tiny
  \centering
  \resizebox{0.48\textwidth}{!}{
  \begin{tabular}{|l|c|c|c|}
    \hline
    \textbf{Model} & \textbf{Architecture} & \textbf{Params} & \textbf{\# Langs} \\
    \hline
    mT5~\cite{xue-etal-2021-mt5}         & ED  & 580M & 101 \\
    mBART50~\cite{tang2020multilingual}     & ED  & 680M  & 50  \\
    XLM-RoBERTa~\cite{conneau-etal-2020-unsupervised} & EO     & 550M  & 100 \\
    Gemma 2 Instruct~\cite{team2024gemma}     & DO  & 9B   & 1  \\
    Gemma 3 1B Instruct~\cite{team2025gemma}     & DO  & 1B   & 35  \\
    Gemma 3 270m Instruct~\cite{team2025gemma}     & DO  & 270m   & 35  \\
    Llama 3.1 Instruct~\cite{patterson2022carbon}   & DO     & 8B    & 8   \\
    Llama 3.2 Instruct~\cite{grattafiori2024llama}   & DO     & 1B    & 8   \\
    \hline
  \end{tabular}}
  \caption{PLMs used in the experiments.  EO - Encoder-Only, DO - Decoder-Only, and ED - Encoder-Decoder}
  \label{tab: PLMs}
\end{table}

\begin{table*}[ht]
\tiny
\centering
\resizebox{\textwidth}{!}{
\begin{tabular}{|>{\centering\arraybackslash}p{1.5cm}|
                >{\centering\arraybackslash}p{3.2cm}|
                >{\centering\arraybackslash}p{0.65cm}|
                >{\centering\arraybackslash}p{0.65cm}|
                >{\centering\arraybackslash}p{0.65cm}|
                >{\centering\arraybackslash}p{2.5cm}|
                >{\centering\arraybackslash}p{1.6cm}|
                >{\centering\arraybackslash}p{3.15cm}|}
\hline
\textbf{Language} & \textbf{Dataset} & \multicolumn{3}{c|}{\textbf{Number of Sentences}} & \textbf{Family} & \textbf{Resource Level} & \textbf{PLMs} \\
\cline{3-5}
& & \textbf{Train} & \textbf{Val.} & \textbf{Test} & & & \\
\hline
Azerbaijani (\AZ) & \citet{localdoc_2024} & 81,176 & 1,000 & 2,000 & Turkic & Low (Category 3) & mT5, XLM-R \\
Bulgarian (\BG) & \citet{klouchek-batista-navarro-2024-bulgarian} & 20,719 & 1,000 & 2,000 & Indo-European (Slavic) & Low (Category 3) & mT5, XLM-R \\
French (\FR) & \citet{rasaboun_spellingcorrectionfrench, fdemelo_french_news} & 5,393 & 500 & 1,000 & Indo-European (Romance) & High (Category 5) & mT5, mBART50, XLM-R, Llama 3.1 \\
Hindi (\HI) & \citet{etoori-etal-2018-automatic} & 160,000 & 1,000 & 2,000 & Indo-European (Indo-Aryan) & Low (Category 3) & mT5, mBART50, XLM-R, Llama 3.1 \\
Korean (\KO) & \citet{vitruv_err_spelling_kor} & 77,000 & 1,000 & 2,000 & Koreanic & High (Category 4) & mT5, mBART50, XLM-R \\
Sinhala (\SI) & \citet{sonnadara2021sinhala,sudesh2022erroff} & 510,706 & 5,282 & 2,037 & Indo-European (Indo-Aryan) & Low (Category 2) & mT5, mBART50, XLM-R \\
Vietnamese (\VI) & \citet{ngo2022combination} & 4,500 & 500 & 1,000 & Austroasiatic & High (Category 4) & mT5, XLM-R \\
Indonesian (\ID) & \citet{10227254} &387000 & - & - & Austronesian & High (Category 4) & mT5, mBART50, XLM-R \\
Turkish (\TK) & \citet{turhan2025leveraging} & 687,008 & 85,871 & 85,824 & Turkic & High (Category 4) & mT5, mBART50, XLM-R \\
\hline
\end{tabular}}
\vspace{2pt}
\begin{minipage}{\textwidth}
\footnotesize
\textsuperscript{*}The Gemma 3 family of models supports over 140 languages, but does not explicitly specify the complete list.
\end{minipage}
\caption{Details of the languages. Resource level is according to~\citet{ranathunga-de-silva-2022-languages}'s language categorization. The final column shows the PLMs that were pre-trained on data from each language}
\label{tab:languages}
\end{table*}

\begin{table*}[!ht]
\tiny
\centering
\resizebox{\textwidth}{!}{
\begin{tabular}{|l|cc|cc|cc|cc|cc|cc|cc|cc|cc|}
\hline
\multirow{2}{*}{\textbf{Model}} 
& \multicolumn{2}{c|}{\textbf{\AZ} (X, T)} 
& \multicolumn{2}{c|}{\textbf{\BG}, (X,T)} 
& \multicolumn{2}{c|}{\textbf{\FR} (X, T, B, L, L1)} 
& \multicolumn{2}{c|}{\textbf{\HI} (X, T, B, L, L1)} 
& \multicolumn{2}{c|}{\textbf{\KO} (X, T, B)} 
& \multicolumn{2}{c|}{\textbf{\SI} (X, T, B)} 
& \multicolumn{2}{c|}{\textbf{\VI} (X, T)}
& \multicolumn{2}{c|}{\textbf{\ID} (X, T, B)} 
& \multicolumn{2}{c|}{\textbf{\TK} (X, T, B)} \\
\cline{2-19}
& \textbf{Det} & \textbf{Corr}
& \textbf{Det} & \textbf{Corr}
& \textbf{Det} & \textbf{Corr}
& \textbf{Det} & \textbf{Corr}
& \textbf{Det} & \textbf{Corr}
& \textbf{Det} & \textbf{Corr}
& \textbf{Det} & \textbf{Corr} 
& \textbf{Det} & \textbf{Corr} 
& \textbf{Det} & \textbf{Corr} \\
\hline

XLM-R (X)
& {24.72} & {16.41}
& {11.10} & {8.24}
& {57.02} & {51.84}
& {74.64} & {72.58}
& {76.16} & {65.24}
& {12.79} & {11.52}
& {25.51} & {25.80} 
& {47.75} & {43.83} 
& {21.68} & {17.28} \\

mT5 (T)
& {8.31} & {0.91}
& {5.92} & {3.13}
& {7.99} & {2.36}
& {76.50} & {64.28}
& {25.81} & {0.93}
& {5.55} & {3.31}
& {12.13} & {5.90} 
& {46.34} & {41.31} 
& {3.24} & {1.27} \\

mBART (B)
& {54.17} & {45.17}
& {21.46} & {18.91}
& {78.81} & {74.75}
& \underline{89.20} & \underline{88.27}
& \underline{96.57} & \underline{95.24}
&  {55.45} & {52.37}
& {70.97} & {66.96} 
& {56.3} & {52.53} 
& {50.07} & {42.5} \\

Llama 3.1 (L)
& \textbf{61.62} & \textbf{59.71}
& \underline{48.44} & \textbf{50.31}
& \underline{87.71} & \underline{86.73}
& {88.33} & {88.25}
& \textbf{97.32} & \textbf{96.40}
& \underline{61.46} & \underline{61.18}
& {70.35} & {67.44} 
& {69.13} & {67.89} 
& {70.38} & {68.06} \\

Llama 3.2 1B (L1)
& {36.81} & {36.22}
& {20.78} & {23.33}
& {77.43} & {75.00}
& {68.20} & {68.26}
& {88.18} & {87.28}
& {54.93} & {55.16}
& {49.80} & {45.50} 
& {53.92} & {51.48} 
& {41.35} & {38.81} \\

Gemma 2 (G)
& \underline{56.01} & \underline{50.66}
& {48.72} & {48.72}
& \textbf{88.19} & \textbf{87.19}
& \textbf{93.30} & \textbf{93.33}
& {96.36} & {95.13}
& \textbf{62.52} & \textbf{62.31}
& \underline{70.68} & \textbf{78.40} 
& {75.96} & {75.13} 
& {77.13} & {75.46} \\

Mixed (with G)
& {54.58} & {48.96}
& \textbf{49.07} & \underline{48.97}
& {87.57} & {86.25}
& {81.81} & {81.08}
& {95.27} & {93.48}
& {33.01} & {30.61}
& \textbf{72.04} & \underline{70.09} 
& {--} & {--}
& {--} & {--}\\

Gemma 3 1B (G1)
& {42.56} & {40.41}
& {30.39} & {31.48}
& {77.46} & {74.34}
& {64.35} & {63.84}
& {93.13} & {91.01}
& {43.54} & {43.09} 
& {54.35} & {50.83} 
& {55.36} & {52.86} 
& {47.69} & {44.88} \\

Gemma 3 270M (Gm)
& {50.14} & {48.57}
& {29.96} & {33.59}
& {72.64} & {69.13}
& {76.67} & {76.43}
& {95.22} & {94.12}
& {57.42} & {58.43}
& {62.96} & {59.37} 
& {49.48} & {47.03} 
& {39.35} & {38.11} \\

\hline

JamSpell
 & {18.33} & {20.15}
 & {42.45} & {42.02}
 & {44.75} & {47.03}
 & {10.65} & {9.99}
 & {13.46} & {18.29}
 & {36.80} & {36.88}
 & {4.30} & {5.16} 
& {39.59} & {37.72} 
& {34.57} & {32.62} \\
 
JamSpell (Full Train Sets)
 & {20.78} & {23.42}
 & {48.64} & {46.82}
 & {45.00} & {47.35}
 & {17.29} & {17.46}
 & {13.46} & {18.29}
 & \textbf{64.94} & \textbf{67.20}
 & {48.74} & {46.97} 
 & {--} & {--}
 & {--} & {--}
 \\

 ASpell
 & {26.17} & {23.65}
 & {3.63} & {3.87}
 & {40.53} & {36.65}
 & {13.87} & {11.00}
 & {--} & {--}
 & {--} & {--}
 & {32.66} & {26.81} 
& {37.55} & {33.90} 
& {13.07} & {7.32}  \\

  HunSpell
 & {--} & {--}
 & {11.43} & {11.16}
 & {39.31} & {36.84}
 & {8.82} & {6.46}
 & {38.70} & {39.46}
 & {36.68} & {35.77}
 & {36.47} & {33.27} 
& {34.10} & {29.43} 
& {33.02} & {26.85} \\

   SymSpell
 & {--} & {--}
 & {14.24} & {13.71}
 & {19.11} & {18.11}
 & {9.10} & {7.82}
 & {14.32} & {15.04}
 & {1.28} & {0.64}
 & {30.43} & {24.54}  
& {25.97} & {20.48} 
& {46.04} & {44.17} \\

\hline
\end{tabular}}
\caption{Performance of PLMs fine-tuned with 5k sentences from each language. For VI, the full dataset of 4,500 sentences was used). For each language, PLMs that include that language are indicated within brackets. Note that although Llama 3.2 and Gemma 3 are said to have been pre-trained on a multilingual dataset, languages coverage details are not publicly available. For each column, \textbf{Bold} indicates the best performance, and \underline{underline} indicates the second-best.}
\label{tab:multi-language-comparision}
\end{table*}

\newcommand{\dashpair}{\multicolumn{1}{c}{\hspace{1em}--\hspace{0.8em}} & 
                      \multicolumn{1}{c|}{\hspace{1em}--\hspace{0.8em}}}
\begin{table*}[!ht]
\tiny
\renewcommand{\arraystretch}{1.1}
\resizebox{\textwidth}{!}{
\centering
\small
\begin{tabularx}{1.2\textwidth}{|l|l|RR|RR|RR|RR|RR|}
\hline
\multirow{2}{*}{\centering\textbf{Lang}} 
& \multirow{2}{*}{\makecell{\textbf{Training Data Set Size}\\ \textbf{Model}}}
& \multicolumn{2}{c|}{\textbf{5000}} 
& \multicolumn{2}{c|}{\textbf{51071}} 
& \multicolumn{2}{c|}{\textbf{127677}} 
& \multicolumn{2}{c|}{\textbf{255353}} 
& \multicolumn{2}{c|}{\textbf{510706}} \\
\cline{2-12}
\small & 
& \textbf{D-F1} & \textbf{C-F0.5}
& \textbf{D-F1} & \textbf{C-F0.5}
& \textbf{D-F1} & \textbf{C-F0.5}
& \textbf{D-F1} & \textbf{C-F0.5}
& \textbf{D-F1} & \textbf{C-F0.5} \\
\hline
\SI 
& XLMR & 12.79 & 11.52 & 24.30 & 22.55 & 46.78 & 45.24 & 52.14 & 50.98 & 64.07 & 63.45 \\
& sinBert & 8.10 & 6.54 & 21.64 & 20.09 & 29.73 & 27.60 & 30.71 & 28.24 & 49.70 & 48.23 \\

& mT5 & 5.55 & 3.31 & \textbf{78.82} & \textbf{77.56} & \textbf{78.93}  & \textbf{78.25} & \underline{76.25} & {75.64} & 77.07 & 76.45 \\

& mBART50 & {55.45} & {52.37} & 68.22 & 66.62 & 73.27 & 72.37 & 71.90 & 71.27 & 74.86 & 74.40 \\

& Llama 3.1 8B & \underline{61.46} & \underline{61.18} & 69.87 & 69.63 & 72.58 & 72.87 & 74.59 & 75.38 & 75.07 & 75.90 \\

& Llama 3.2 1B & 54.93 & 55.16 & 60.41 & 60.70 & {69.84} & {69.62} & {72.71}& {72.87} & 72.16 & 71.83 \\

& Gemma 2 9B & \textbf{62.52} & \textbf{62.31} & 72.49 & 72.70 & \underline{76.62} & \underline{76.91}& \textbf{78.65} & \textbf{79.01} & \textbf{81.93} & \textbf{82.87} \\

& Gemma 3 1B - LoRa & 43.54 & 43.09 & {60.51}  & {60.93} & {67.28} & {67.67} & {69.47}& {69.82} & 73.05 & 74.31\\

& Gemma 3 1B - Full & {60.28} & {61.06} & \underline{75.34} & \underline{75.49} & {76.19} & {76.46} &  \dashpair & \dashpair \\

& Gemma 3 270m & {57.42} & {58.43} & {70.31} & {70.34} & {73.70} & {73.59} & 75.69  & \underline{76.03} & \underline{78.41} & \underline{78.58} \\
& JamSpell & {38.73} & {36.88} & {55.10} & {56.70} & {60.33} & {62.23} & {62.75} & {65.01} & {64.94}& {67.20} \\
\hline
%\end{tabular}
%\hline
\HI & XLM-R & 74.64 & 72.58 &  88.21 & 87.18 &  96.31 & 96.07 & \dashpair & \dashpair \\
& mT5 & 76.50 & 64.28 &  84.13 & 82.36 &  93.60 & 91.37 & \dashpair & \dashpair \\
& mBART & \underline{89.20} & \underline{88.27} &  95.93 & 95.40 & \textbf{97.99} & {97.59} & \dashpair & \dashpair \\
& Llama 3.1 & 88.33 & 88.25 &  \textbf{97.13} & \textbf{97.16} &  {97.75} & \textbf{97.82} & \dashpair & \dashpair \\
& Lllama 3.2 1B & {68.20} & {68.26} &  {93.83} & {93.98} &  {93.83} & {93.98} & \dashpair & \dashpair \\
& Gemma 2 & \textbf{93.30} & \textbf{93.33} &  \underline{96.59} & \underline{96.61} &  \underline{97.76} & \underline{97.80} & \dashpair & \dashpair \\
& Gemma 3 1B & {64.35} & {63.84} &  {88.02}& {87.97} &  {92.73} & {92.70} & \dashpair & \dashpair \\
& Gemma 3 270m & {76.67}& {76.43} &  {90.36} & {90.26} &  93.37 & 93.24 & \dashpair & \dashpair \\
& JamSpell & {10.82}& {10.00} &  {13.30} & {13.28} &  17.29 & 17.46 & \dashpair & \dashpair \\
\hline
\end{tabularx}
}
\caption{Performance across different dataset sizes for Sinhala and Hindi. Each dataset size corresponds to approximately 1\%, 10\%, 25\%, 50\%, 100\% of the Sinhala training dataset.}
\label{tab:SiHi-dataset-size-comparision}
\end{table*}

\section{Evaluation}

JamSpell\footnote{\url{https://github.com/bakwc/JamSpell}} and ASpell\footnote{\url{http://aspell.net/}} serve as the rule-based baseline spell correctors. For JamSpell, we trained a custom binary file per language using the same dataset used to train the PLMs. This provided an alphabet file for each language. ASpell does not need such a corpus. Instead, it relies on a language-specific dictionary. However, Sinhala and Korean were excluded from the ASpell implementation due to specific technical constraints: Sinhala lacks official dictionary support and requires complex legacy configuration for Unicode compatibility, while Korean’s logographic roots and highly agglutinative morphology exceed the capabilities of ASpell’s affix-based architecture. Similarly, we evaluated Hunspell using language-specific dictionaries and morphological rules with edit-distance-based candidate generation ranked by phonetic similarity, and SymSpell using \textit{Symmetric Delete Graphs} with edit distance thresholds of 1 and 2 for fast candidate retrieval. However, both systems lacked native Azerbaijani language support, limiting their evaluation on this language. Further implementation details are in Appendix~\ref{sec: rule-based}.

PLMs were trained with 5k sentences for each language. Results are in Table~\ref{tab:multi-language-comparision}. Results under the ‘Mixed’ column were produced by fine-tuning
Gemma 2 by combining 5k from all languages. We were able to get results for JamSpell with the complete dataset for some of the languages, but its not considere for the discussion. All PLMs, except mT5 and XLM-R for some languages, outperform JamSpell (except for \BG ~in some PLMs) and ASpell. In particular, Gemma 3 270M, despite being fine-tuned with just 5k sentences, outperforms ASpell in all experiments and JamSpell in all but \BG. However, XLM-R and mT5 results are rather surprising - despite being pre-trained with all the languages that we considered, they showed the worst performance, except for \HI. Size of the LLM does have a positive impact on the results. Llama 3.1 8B consistently outperforms the Llama 3.2 1B by a significant margin. Similarly, Gemma 2 outperforms its smaller siblings. However, Gemma 3 1B does not seem to always outperform Gemma 3 270M model. This is due to LoRA-based fine-tuning (see Section~\ref{sec:experiment_setup}). In order to investigate the impact of LoRA-based fine-tuning, we conducted several full-fine-tuning experiments with Gemma 3 1B. As show in Table~\ref{tab:SiHi-dataset-size-comparision}, when Gemma 3 1B is fully fine-tuned, it surpasses the smaller 270M model. 

We also evaluated the impact of dataset size, using ~\SI ~and ~\HI. As per Table~\ref{tab:SiHi-dataset-size-comparision}, for \SI, when the full dataset is used, Gemma 2 9B shows the best result, followed by Gemma 3 270m. Despite having a very low result with 5k sentences, mT5 results pick up when the dataset size increases, and shows the third-best result for the full training set. As for \HI, Gemma 2, Llama 3.1 and mBART50 fine-tuned with the full dataset show similar results. \citet{10883152} states that mT5 models require sufficient data to perform effectively; hence, it is reasonable to assume that the insufficiency of the fine-tuning dataset caused the poor results for mT5 in this experiment. However, for Hindi, mT5 performed significantly better than the other models, which could be due to the abundance of Hindi in the pre-training data for mT5.

\iffalse
\begin{table*}[h!]
\centering
\small
\begin{tabular}{|l|l|*{7}{p{1.2cm}|}}
\hline
\textbf{Model} & \textbf{Training Set} & \textbf{D-F1} & \textbf{C-F0.5} & \textbf{In-Out} & \textbf{In-Out=Exp} & \textbf{In-Out!=Exp} & \textbf{(In=Exp) !=Out} & \textbf{Out=Exp} \\
\hline
Gemma 2 9B & 100\% & 81.93 & 82.87 & 789 & 770 & 19 & 264 & 1334 \\
Llama 3 8B & 100\% & 75.07 & 75.90 & 660 & 642 & 18 & 392 & 1240 \\
Gemma 2 9B & 5000 SAD & 66.89 & 69.12 & 1325 & 971 & 354 & 63 & 1099 \\
Llama 3 8B & 5000 SAD & 66.73 & 68.41 & 1337 & 989 & 348 & 45 & 1097 \\
Gemma 2 9B & 5000 & 62.52 & 62.31 & 822 & 712 & 110 & 322 & 939 \\
Llama 3 8B & 5000 & 61.46 & 61.18 & 824 & 715 & 109 & 319 & 930 \\
\hline
\end{tabular}
\caption{Model performance and error analysis metrics for Sinhala. Comparing models finetuned using 100\% training set vs 5000 data points vs 5000 data points with only substitution, deletion and addition errors. \nds{I don't understand what is happening in the table headings here and I cannot find this to be refereed in the body text either.}}
\label{tab:model_analysis}
\end{table*}
\fi

\section{Case Study - Sinhala Spell Correction}

Sinhala is an Indo-Aryan language, which is only being used by a population of about 20 million in the island nation of Sri Lanka. According to ~\citet{ranathunga-de-silva-2022-languages}'s language categorisation, Sinhala is a low-resource language.  

As an attempt to improve LLM performance for spell correction, we experimented with in-context learning and RAG~\cite{lewis2020retrieval}. Table~\ref{tab:zero-few-shot-combined} presents the results of the zero-shot and few-shot (n=4) results of the un-finetuned (B) version  of  the LLMs, as well as their fine-tuned (F) version (with the full Sinhala training set). The fine-tuned version significantly outperforms its counterpart. Few-shot prompting further boosts Llama 3.1 results albeit a slight drop in Gemma 2 results. 

\begin{table}[thb]
\small
\centering
\setlength{\tabcolsep}{6pt}
\renewcommand{\arraystretch}{1.2}
%\begin{adjustbox}{width=\columnwidth}
\resizebox{0.4\textwidth}{!}{
\begin{tabular}{|l|c|c|c|c|}
\hline
\textbf{Exp} 
& \multicolumn{2}{c|}{\textbf{Llama 3.1}} 
& \multicolumn{2}{c|}{\textbf{Gemma 2}} \\
\cline{2-5}
& \textbf{D-F1} & \textbf{C-F0.5} 
& \textbf{D-F1} & \textbf{C-F0.5} \\
\hline
ZS (B)    & {51.13}         & {38.58}        & 4.17           & 4.06 \\
FS (B)    & {48.73}         & {49.73}        & 14.20          & 13.03 \\
ZS (F)    & 75.07           & 75.90          & \textbf{81.93} & \textbf{82.87} \\
FS (F)    & \textbf{79.66}  & \textbf{80.67} & 80.69          & 82.04 \\
\hline
RAG 1     & -               & -              & 79.32          & 80.87 \\
RAG 2     & -               & -              & 80.53          & 81.07 \\
\hline
\end{tabular}
}
%\end{adjustbox}
\vspace{0.5em}
\captionof{table}{Zero-shot (ZS), Few-shot (FS) and RAG results for the un-finetuned (B) and fine-tuned (F) LLMs.}
\label{tab:zero-few-shot-combined}
\end{table}

\citet{guo-etal-2024-retrieval} and \citet{dong-etal-2025-retrieval} employed RAG to improve LLM performance for spell correction. In \citet{guo-etal-2024-retrieval} (RAG 1), a candidate correction word is generated for each word in the input sentence. Then these candidates are provided as context to the prompt during inference. In \citet{dong-etal-2025-retrieval}  (RAG 2), for each input sentence, three contextually similar sentences retrieved from a vector store (built on a separate dataset~\cite{aravinda2025sinllama})  were used as contextual input to the prompt during inference. We employed these RAG techniques with Gemma 2 9B Model fine-tuned with the full Sinhala dataset (henceforth called \texttt{Gemma-2-full}), using a Faiss vector database \citep{douze2024faiss}. The candidate error list provided in \citet{sudesh2022erroff} was used for the first experiment, while the public dataset from \citet{aravinda2025sinllama} was used for the second. Table \ref{tab:zero-few-shot-combined} shows the results. These techniques did not give results similar to those reported for English and Chinese by previous research. This could be due to the limited effectiveness of vector retrieval in Sinhala, where contextual similarity retrieval is less reliable due to underdeveloped embedding spaces~\cite{nzeyimana2025kinyacolbert}. Noise in the datasets could be another reason. This issue should be further investigated by future research.
\begin{table}[thb]
  \centering
  \small
  \begin{tabular}{|l|c|c|}
\hline
\textbf{Domain} & \textbf{D-F1} & \textbf{C-F0.5} \\

\hline
Government & 34.50 & 33.60 \\
Newspaper & 57.58 & 58.33 \\
Magazine  & 44.23 & 44.23 \\
Socialmedia & 31.15 & 31.71 \\
Wikipedia  & 36.36 & 36.36 \\
\hline
\end{tabular}
\caption{Domain-specific results for \texttt{Gemma-2-full}}
\label{tab:domain_results}
\end{table}

\begin{table}[h!]
\centering
\small
\renewcommand{\arraystretch}{1.2}
\resizebox{0.48\textwidth}{!}{
\begin{tabular}{|l|c|c|c|c|}
\hline
\multirow{2}{*}{\textbf{Model}} 
& \multicolumn{2}{c|}{\textbf{Original Set}} 
& \multicolumn{2}{c|}{\textbf{Synthetic Set}} \\
\cline{2-5}
& \textbf{D-F1} & \textbf{C-F0.5}
& \textbf{D-F1} & \textbf{C-F0.5} \\
\hline
mT5 & 5.55 & 3.31 & 1.68 & 0.76 \\
mBART50 & {55.45} & {52.37} & {49.61} & {47.30} \\
Gemma 2 & 62.52 & 62.31 & {66.89} & {69.12} \\
Llama 3.1 & 61.46 & 61.18 & {66.73} & {68.41} \\
\hline
\end{tabular}
}
%\end{adjustbox}
\caption{Original test set evaluated with models finetuned with original and synthetic error data. }
\label{tab:different-error-results}
\end{table}

\begin{table}[h!]
\centering
\small
\renewcommand{\arraystretch}{1.2}
\resizebox{0.48\textwidth}{!}{
\begin{tabular}{|l|l|c|c|c|c|}
\hline
\makecell{ \textbf{Train Set}} & \textbf{Test Set} & \textbf{Accuracy} & \textbf{Precision} & \textbf{Recall} & \textbf{F1} \\
\hline
Original & Original  & 0.81 & 0.87 & 0.81 & 0.78 \\
Corrected & Original  & 0.92 & 0.93 & 0.92 & 0.92 \\
Corrected & Corrected & \textbf{0.93} & \textbf{0.94} & \textbf{0.93} & \textbf{0.93} \\
\hline
\end{tabular}
}
\caption{Finetuning Performance of SinLlama~\cite{aravinda2025sinllama} on Sinhala text classification using datasets spell corrected using \texttt{Gemma-2-full}.}
\label{tab:model_performance}
\end{table}

In order to test the robustness of LLMs, we tested \texttt{Gemma-2-full} with data from different domains that~\citet{sonnadara2021sinhala} provided. According to Table~\ref{tab:domain_results}, domain has a significant influence on the LLM's spell correction ability. Social media data is the most challenging, possibly due to the code-mixed and informal writing style of such text. This is further explained by the results in Table~\ref{tab:error-percentage-comparison-transposed} of Appendix~\ref{sec:data_creation}- when the amount of errors in the test set increases, LLM accuracy drops. As expected, the performance of the other PLMs drops substantially, with the notable exception of mT5, which remarkably maintains performance comparable to that of the LLMs.

According to the publicly available dataset details (see Appendix~\ref{app:lang}), datasets we used were created by incorporating language-specific spelling errors. One could see this as a limiting factor. Therefore we investigated the impact of using synthetic data to train PLMs for the task of spell correction. We started with the original Sinhala 5000 sentences without spelling errors and added addition, substitution, deletion errors in a similar percentage to that of~\citet{sonnadara2021sinhala} and tested with the original test set. As shown in Table~\ref{tab:different-error-results}, synthetic data has a detrimental effect on mBART and mT5. However, for both Llama and Gemma, it is the opposite - both LLMs report gains. We hypothesize that using synthetic errors retains language-specific spelling details in the corpus, which helps the LLM to learn spelling rules of the language.

In order to show the usefulness of spell correction on downstream tasks, we spell corrected a Sinhala sentiment analysis dataset ~\cite{senevirathne2020sentiment} using the \texttt{Gemma-2-full} model. As shown in Table~\ref{tab:model_performance}, simply spell correcting the test set resulted in an F1 gain of 14\%.

\section{Conclusion}
In this paper, we carried out an empirical evaluation on the effectiveness of different PLMs for spell correction. We show that for most PLMs, just 5k training data is more than sufficient to beat rule-based spell checkers. For larger LLMs, even synthetic data is sufficient. Our experiments with Sinhala as a case study highlight the challenges and benefits of spell correction for LRLs.

\section{Acknowledgments}
We thank Emojot (Pvt) Ltd for providing the computing resources. We also thank Haritha Gunarathna,Vihanga Jayawickrama and Thamali Wijewardhana for their support at various stages of the project. 

\section*{Limitations}
Due to the limitations in computing resources, our experiments considered only 8 PLMs. We could not go beyond 9B for LLM experiments due to the same reason. We could only find 9 spell correction datasets. Dataset size-based experiments had to be limited to only Sinhala and Hindi, due to computer resource limitations. 
\section*{Ethical Considerations}
We use existing datasets from Huggingface. Whatever the biases in these datasets, as well as PLMs may reflect on the spell corrector output.
\small

\bibliography{anthology, latex/references, latex/custom}

\appendix
 
\section{\LMS{} Toolkit}
 \label{sec:lmspell}

 \begin{figure*}[htbp]
    \centering
    \includegraphics[width=0.8\textwidth]{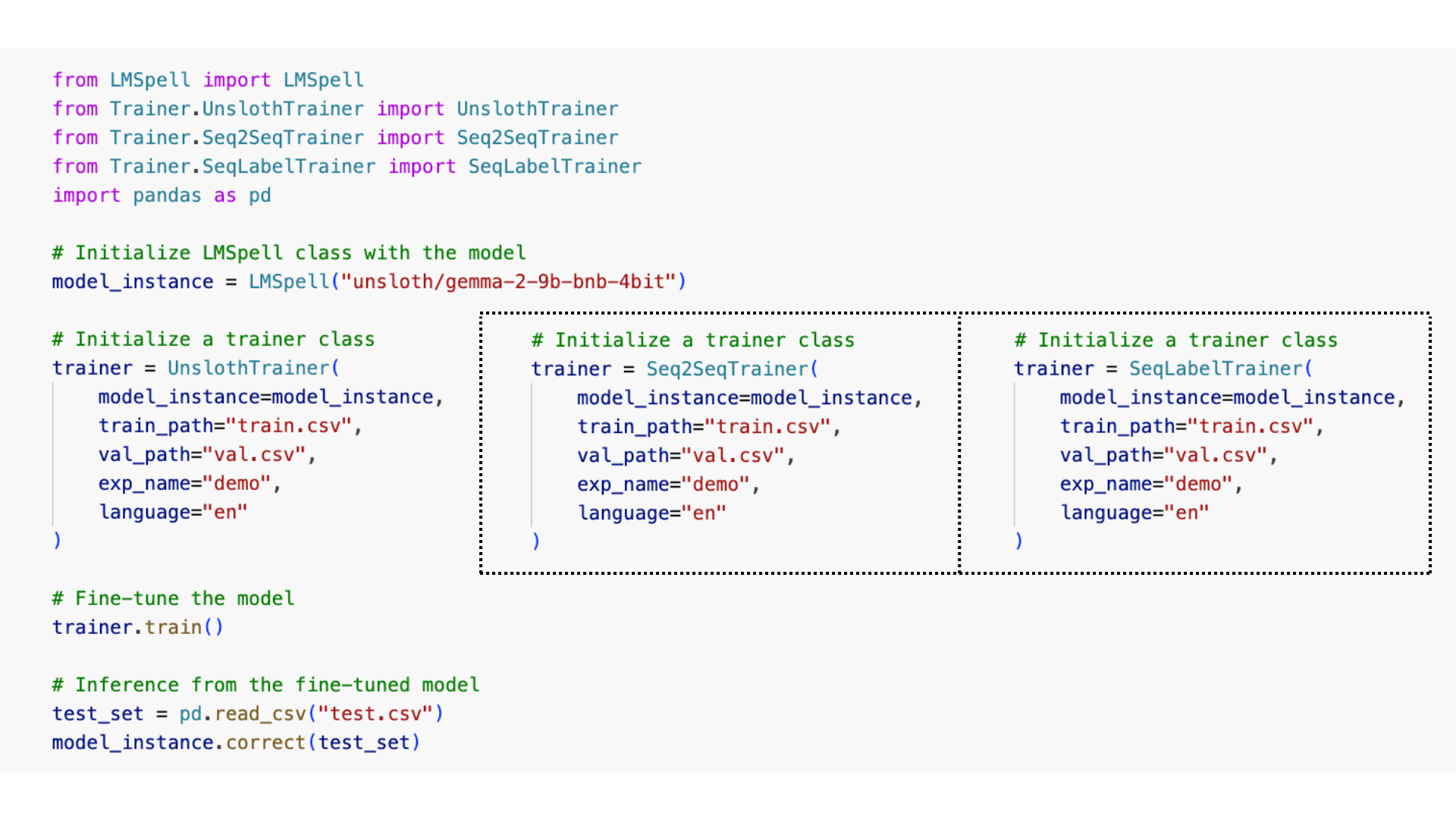}
    \caption{Interface of \LMS{}}
    \label{fig: LMSpell Toolkit}
\end{figure*}

\begin{figure}[htbp]
  \centering
  \includegraphics[width=0.9\columnwidth,page=1]{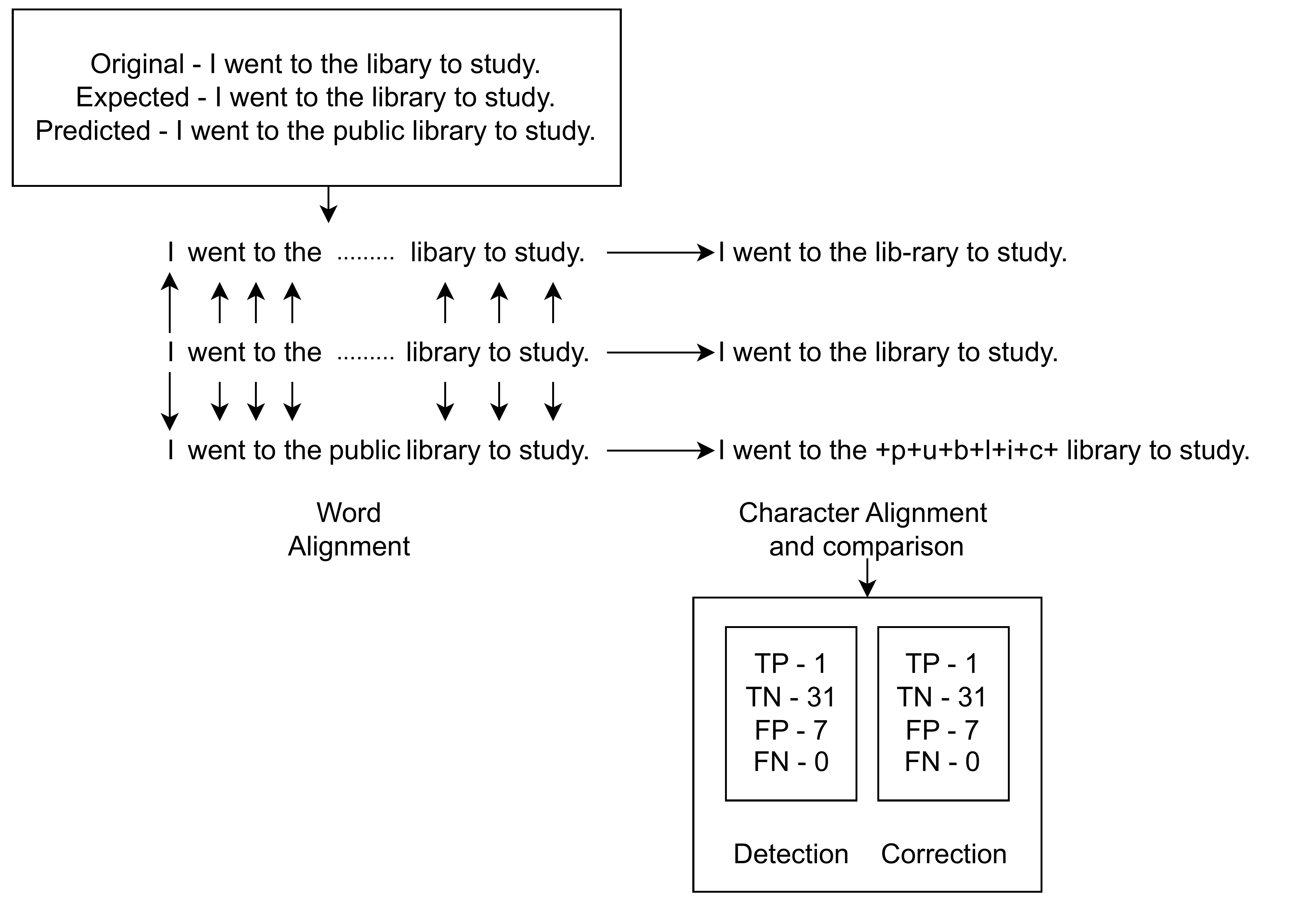}
  \caption{Evaluation workflow during hallucination}
  \label{fig:hallucination-demo-pdf}
\end{figure}

 For encoder-based and encoder-decoder-based models, the model pipeline integrates Accelerate~\citep{accelerate} with DeepSpeed~\citep{rasley2020deepspeed} for efficient fine-tuning. Unsloth~\citep{unsloth}, LoRA~\citep{hu2021loralowrankadaptationlarge} implementation from PEFT~\citep{peft} and SFTTrainer~\citep{ouyang2022training} are used to fine-tune decoder-based models. Figure~\ref{fig: LMSpell Toolkit} shows the programmable interface of \LMS{}. The prompt used for decoder models is shown below. Figure ~\ref{fig:hallucination-demo-pdf} shows how evaluation works in our library.
\paragraph{Handling Hallucination:}
 In the context of spell correction, hallucination refers to cases where a PLM introduces content that is neither in the original nor the expected sentences. As illustrated in Figure~\ref{fig:hallucination-demo-pdf}, the model has inserted the adjective \textit{public} to the predicted sentence. Even though the predicted sentence has correct spelling, this is treated as an incorrect output in this context, compared to the original sentence. These hallucinations shift the word alignment of the predicted sentence compared to the original and the expected sentences. This must be considered when evaluating spell-checkers that use PLMs. Previous research that has its code publicly available has not solved this issue successfully. For instance, the evaluation scripts used by \citet{sonnadara2021sinhala} and \citet{jayanthi-etal-2020-neuspell} lacked proper string alignment, making them unfit for evaluating hallucinated outputs. Without proper alignment, recall, precision and F-scores can be significantly distorted, especially when hallucinations result in insertions and deletions that alter the sentence structure.

To address this, a new evaluation script was developed. As shown in Figure~\ref{fig:hallucination-demo-pdf}, first it aligns the original, predicted and expected sentences at the word level, identifying any hallucinated words in the predicted sentence and treating them as false positives. After flagging those words in the predicted sentence, the remaining words are aligned on a character level to compare with each other. Character-level evaluation was identified as more appropriate, since PLMs process text at the character-level.
\begin{tcolorbox}
\textit{You are an expert \{language\} spell corrector. Below is a sentence in \{language\} language. It may or may not have spelling mistakes. Give the corrected output in \{language\}.}
\end{tcolorbox}

\begin{table}[h!]
\centering
\renewcommand{\arraystretch}{1.2}
\begin{adjustbox}{width=\columnwidth}
\begin{tabular}{|l|c|c|c|c|c|c|}
\hline
\multirow{2}{*}{\textbf{Model}} 
& \multicolumn{2}{c|}{\textbf{Original (7\%)}} 
& \multicolumn{2}{c|}{\textbf{41\%}} 
& \multicolumn{2}{c|}{\textbf{65\%}} \\
\cline{2-7}
& \textbf{D-F1} & \textbf{C-F0.5}
& \textbf{D-F1} & \textbf{C-F0.5}
& \textbf{D-F1} & \textbf{C-F0.5} \\
\hline
Sinbert
& {49.70} & {48.23} 
&  41.85 & 36.24
&  37.43 & 29.63\\
mT5 
& {77.07} & {76.45} 
& 58.02 & 51.98
& 50.14 & 40.63\\
mBart
& {74.86} & {74.40} 
& 28.81 & 20.44
& 21.00 & 10.33 \\
Gemma 2 
& \textbf{81.93} & \textbf{82.87} 
& 53.25 & 60.95 
& 45.45 & 55.33 \\
Llama 3.1 
& 78.57 & 79.47 
& \textbf{61.76} & \textbf{70.05} 
& \textbf{53.03} & \textbf{63.80} \\
\hline
\end{tabular}
\end{adjustbox}
\captionof{table}{Performance for different error percentages of the test set, using PLMs fine-tuned with the full training set. Best PLM for each set is highlighted.}
\label{tab:error-percentage-comparison-transposed}
\end{table}

\section{Experiment Details}
\label{sec:hyper-params}

\begin{table}[h!tb]
\small
  \centering
\resizebox{0.48\textwidth}{!}{
  \begin{tabular}{|l|c|c|c|c|c|}
    \hline
    \textbf{Parameter} & \textbf{mT5} & \textbf{mBART50} & \textbf{XLM-R} & \textbf{Sinbert} \\
    \hline
    Batch Size                 & 16     & 16     & 16     & 32    \\
    Mixed Precision            & bf16   & fp16   & fp16   & fp16    \\
    Zero Stage                 & 2      & 2      & 2      & 2    \\
    Max Seq Length (Train)     & 128    & 128    & 128    & 128    \\
    Patience                   & 5      & 5      & 5      & 5     \\
    ZWJ Fix                    & Yes    & Yes    & Yes    & No   \\
    Grad. Acc. Steps           & 4      & 4      & 4      & 2     \\
    Initial Learning Rate      & 1e-5   & 1e-5   & 1e-5   & 1e-5  \\
    Learning Scheduler         & Linear + Warmup     &  Linear + Warmup   &  Linear + Warmup &  Linear + Warmup     \\
    No. of Warmup Steps          & 10     & 10     & 10 & 10    \\
    \hline
  \end{tabular}
  }
  \caption{Training parameters for Encoder-based and Encoder-Decoder models.}
  \label{tab:training_parameters_1}
\end{table}

\begin{table}[h!tb]
% \tiny
\renewcommand{\arraystretch}{1.2}
  \centering
  \resizebox{\columnwidth}{!}{
  \begin{tabular}{|l|c|c|c|c|c|}
    \hline
    \textbf{Parameter} & \textbf{Gemma 2 9b} & \textbf{Gemma 3 1b} & \textbf{Gemma 3 270m} & \textbf{Llama 3.1 8b} & \textbf{Llama 3.2 1b} 
    \\
    \hline
    Batch Size                 & 4    & 8    & 4   & 4 & 8\\ 
    ZWJ Fix                    & No   & No   & No  & No & No\\ 
    Grad. Acc. Steps           & 2    & 1    & 1   & 2 & 1 \\
    Intial Lr. Rate            & 1e-5 & 1e-5   & 1e-5  & 1e-5 & 1e-5\\
    R                          & 8    & 16   & None\tablefootnote{Full Model is finetuned instead of using LORA} & 8 & 16 \\
    Flash Attention            & Yes  & Yes  & No & Yes & Yes\\
    Optimizer                  & Adam & Adam & Adam & Adam & Adam\\
    \hline
  \end{tabular}}
  \caption{Training parameters for LLMs.}
  \label{tab:training_parameters_2}
\end{table}

Details on the PLMs are in Table~\ref{tab: PLMs}, along with hyper-parameter details in Tables~\ref{tab:training_parameters_1} and~\ref{tab:training_parameters_2}. We picked a representative mixture of \textit{Encoder-Only}, \textit{Decoder-Only}, and \textit{Encoder-Decoder} models for our experiments. For inference, default values were used for all models.  

%\onecolumn
\section{Languages Used}
\label{app:lang}

Table~\ref{tab:languages} shows the details of the languages used in this study. Given the wide-spread nature of the Indo-European (IE) language family, we have opted to also show in parenthesis the subtree to which IE languages belong. We have attempted to get a good spread of languages where they cover the range of Category 2 up to Category 5 by the definition provided by~\citet{ranathunga-de-silva-2022-languages}. Due to the even lower resource nature of the languages, we were unable to find any suitable datasets from Category 0 or Category 1.  Dataset descriptions are given below. Note that not all datasets are associated with a dataset description. We could only find details for Bulgrian and Vietnamese.

\textbf{Bulgarian} (\cite{klouchek-batista-navarro-2024-bulgarian})  - The dataset was derived from Bulgarian Wikipedia articles. To filter out noisy content such as titles, links, and fragments, only sentences containing at least three words and at least one verb were retained. The dataset was then constructed by automatically introducing spelling errors into otherwise correct sentences using Python scripts. For each selected sentence, the original form was kept as the reference, an error of a predefined type was induced whenever the sentence satisfied the relevant linguistic conditions, and the correct and erroneous versions were paired. Errors were applied only to lemmas longer than three characters and excluding capitalized words, to avoid function words and probable named entities. The resulting data covers seven error types: vowel change, double consonant alternation, end-of-lemma consonant alternation, deletion of repeated \textit{t} or \textit{n}, loss of \textit{t} or \textit{d} sounds, random character substitution, and semantic errors that preserve syntactic validity while disrupting sentence meaning.

\textbf{Vietnamese} (\cite{ngo2022combination}) - The dataset has been sourced from a corpus that contains articles from multiple news outlets. After pre-processing the dataset, common spelling error types such as fat finger, region-based errors and telex errors (since Viatnames use a derivation of Latin script) have been randomly introduced to the dataset. 
\setlength{\tabcolsep}{3pt}

\section{Computational Resources}
\label{sec:computational_resources}
%\small
All experiments were conducted on cloud infrastructure, primarily using Google Cloud Platform (GCP) and Kaggle with NVIDIA T4 and L4 GPUs. Each result was obtained by averaging the outcomes from three experiments. The GPU hours used for fine-tuning the full Sinhala dataset are shown in Table~\ref{tab:finetuning-times}.  For testing, the full test set was always used, and the GPU hours are reported in Table~\ref{tab:testing-times}.

\begin{table}[ht]
\renewcommand{\arraystretch}{1.}
\small
 \centering
  \begin{tabular}{|l|c|c|c|c|}
\hline
\textbf{Model} & \textbf{GPU Type} & \textbf{Time (hours)} \\
\hline
sinBert & NVIDIA T4 (x4)& 8 \\
XLM-R & NVIDIA T4 (x4)& 22 \\
mT5 & NVIDIA T4 (x4)& 38 \\
mBART & NVIDIA T4 (x4)& 22 \\
Gemma 2 & NVIDIA L4 (x1) & 70 \\
Gemma 3 270M & NVIDIA T4 (x1) & 12 \\
Gemma 3 1B & NVIDIA L4 (x1) & 20 \\
LLaMA 3.1 & NVIDIA L4 (x1) & 64 \\
LLaMA 3.2 1B & NVIDIA L4 (x1) & 20 \\
\hline
\end{tabular}
\caption{Training time for Sinhala Full dataset}
\label{tab:finetuning-times}
\end{table}

%\filler{6}

\begin{table}[ht]
\renewcommand{\arraystretch}{1}
\small
  \centering
  \begin{tabular}{|l|c|c|c|c|}
\hline
\textbf{Model} & \textbf{GPU Type} & \textbf{Time (hours)} \\
\hline
sinBert & NVIDIA T4 (x2)& 0.2 \\
XLM-R & NVIDIA T4 (x2)& 0.5 \\
mT5 & NVIDIA T4 (x2)& 0.5 \\
mBART & NVIDIA T4 (x2)& 0.5 \\
Gemma 2 & NVIDIA T4 (x1)& 4 \\
Gemma 3 1B & NVIDIA T4 (x1) & 1 \\
Gemma 3 1B & NVIDIA T4 (x1) & 1.5 \\
LLaMA 3.1 & NVIDIA T4 (x1)& 6 \\
LLaMA 3.2 1B & NVIDIA T4 (x1) & 1.5 \\
\hline
\end{tabular}
\caption{Testing time for Sinhala Full dataset}
\label{tab:testing-times}
\end{table}

\section{Error Visual Examples}
\label{appendix:error-categories}

% \phantomsection
% \addcontentsline{toc}{section}{Error Categories and Visual Examples}

% \begin{table}[ht]
% \tiny
% \centering
% \begin{tabular}{|l|c|c|}
% \hline
% \textbf{Error Type} & \textbf{Percentage} \\
% \hline
% Deletion & 42.1 \\
% Insertion (Consonant) & 24.6 \\
% Proximity\tablefootnote{The most popular Sinhala keyboard layout which is the Helakuru Phonetic layout, was considered for this.} & 20.5 \\
% Similar shape & 4.3 \\
% Diacritic & 3.5 \\
% Prenasalized & 2.2 \\
% na-Na-la & 1.9 \\
% Aspirated and unaspirated consonant & 0.6 \\
% Retroflex and palatal sibilant & 0.2 \\
% \hline
% \end{tabular}
% \caption{Error percentages in Sinhala text}
% \label{tab:error-percentages-in-sinhala}
% \end{table}

\begin{minipage}{\linewidth}
\begin{itemize}[itemsep=0pt, parsep=0pt, topsep=0pt]
    \item Insertion
    \begin{itemize}[itemsep=0pt, parsep=0pt, topsep=0pt]
        \item Consonant
        \item Diacritic
    \end{itemize}
    \item Deletion
    \item Substitution
    \begin{itemize}[itemsep=0pt, parsep=0pt, topsep=0pt]
        \item Diacritic \quad \raisebox{-0.6ex}{\includegraphics[height=2.5ex]{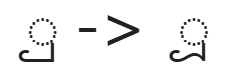}}
        \item Similar shape \quad \raisebox{-0.6ex}{\includegraphics[height=2.5ex]{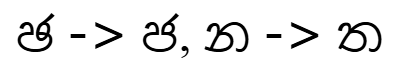}}
        \item Prenasalized \quad \raisebox{-0.6ex}{\includegraphics[height=2.5ex]{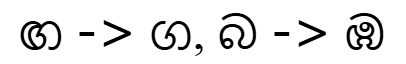}}
        \item na-Na-la \quad \raisebox{-0.6ex}{\includegraphics[height=2.5ex]{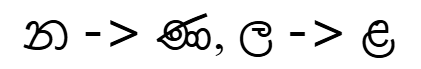}}
        \item Aspirated and unaspirated consonant \quad \raisebox{-0.6ex}{\includegraphics[height=2.6ex]{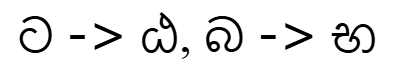}}
        \item Retroflex and palatal sibilant \quad \raisebox{-0.6ex}{\includegraphics[height=2.5ex]{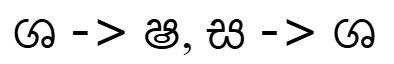}}
        \item Encoding \quad \raisebox{-0.6ex}{\includegraphics[height=2.5ex]{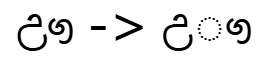}}
        \item Other \quad \raisebox{-0.6ex}{\includegraphics[height=2.5ex]{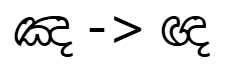}}
    \end{itemize}
\end{itemize}
\end{minipage}

\section{Rule-Based Spell Correctors}
\label{sec: rule-based}
To support all seven target languages, we trained custom binary models by aggregating the ground-truth text from each training dataset into language-specific corpora and defining corresponding alphabet files. The ASpell implementation utilises a Python wrapper to interface with the GNU Aspell engine via \texttt{subprocess} calls, leveraging UTF-8 encoding for multilingual script support. 

In contrast, JamSpell is integrated via a custom pipeline that builds the library from source with SWIG\footnote{\url{https://www.swig.org/}}-generated Python bindings. To support all seven target languages, we trained custom binary models by aggregating the ground-truth text from each training dataset into language-specific corpora and defining corresponding alphabet files. These custom models were then utilised through the \texttt{TSpellCorrector} class to perform context-aware, row-by-row fragment correction across the test datasets.

%We evaluated Hunspell and SymSpell across nine multilingual datasets using a standardised evaluation framework. 
For Hunspell, we utilised the default spell-checking engine with language-specific dictionaries and morphological rules, performing direct dictionary lookups followed by edit-distance-based candidate generation when words were not found. Corrections were ranked by phonetic similarity and keyboard proximity scores provided by the Hunspell library. For SymSpell, we constructed Symmetric Delete Graphs from the same language dictionaries with edit distance thresholds of 1 and 2, enabling fast lookup of correction candidates during the evaluation phase through precomputed deletion variants. %Both systems generated predictions for the noisy text column in each test dataset, and the corrections were evaluated against gold-standard reference texts.

\section{Process of Creating Test Data with different Error Percentages}
\label{sec:data_creation}

 We define error percentage as the proportion of characters in the dataset that is different compared to the correct sentence. The original test set contained ~7\% naturally occurring errors. To evaluate model robustness, we applied our error injection logic at different intensities. We calculated the distribution of different error types in Sinhala text using two distinct corpora: one is a spell-corrected corpus composed of government documents~\cite{fernando2020data, ranathunga2018si} and textbooks, and the other is an uncorrected corpus representing real-world text~\cite{kudugunta2023madlad}. Based on these distributions, we injected new errors following the observed proportions, with the error intensity controlled by adjusting the pass-through rate (the likelihood that a word remains unchanged). This resulted in test sets with approximately 41\% and 65\% error rates. Note that these percentages were not manually chosen but reflect the empirical outcome of applying different levels of corruption.

%\filler{8-10}

\end{document}